\documentclass[10pt,twocolumn,letterpaper]{article}

\usepackage{multirow, booktabs}
\usepackage[table]{xcolor}
\usepackage{cvpr}              

\usepackage{url}            
\usepackage{xcolor}         

\usepackage{times}
\usepackage{microtype} 
\usepackage{graphicx} 
\usepackage{wrapfig}
\usepackage{balance} 

\usepackage{algorithm}
\usepackage{algorithmic}

\usepackage{helvet}
\usepackage{courier}

\usepackage{makecell}
\usepackage{graphicx}
\usepackage{color}
\usepackage{amsfonts}
\usepackage{amsmath}
\usepackage{amssymb}

\usepackage{bm}

\usepackage{multirow}
\usepackage{booktabs}

\def\ie{\mbox{\textit{i.e.}, }}
\def\eg{\mbox{\textit{e.g.}, }}
\def\wrt{\mbox{\textit{w.r.t. }}}

\def\mI{{\mathcal I}}

\def\mU{{\mathcal U}}

\DeclareMathAlphabet\mathbfcal{OMS}{cmsy}{b}{n}

\def\0{{\bf 0}}
\def\1{{\bf 1}}

\def\bF{{\bm{F}}}
\def\bG{{\bm{G}}}

\def\bI{{\bm{I}}}

\def\bK{{\bm{K}}}

\def\bQ{{\bm{Q}}}

\def\bV{{\bm{V}}}

\def\bZ{{\bm{Z}}}

\def\br{{\bm r}}
\def\bs{{\bm s}}

\def\bx{{\bm x}}

\usepackage{ntheorem}

\newtheorem*{*thm}{Theorem}

\newtheorem*{*lemma}{Lemma}

\usepackage{stackengine}
\usepackage{tabularx}
\usepackage{threeparttable}
\usepackage{arydshln}

\definecolor{mygray}{gray}{.9}

\usepackage{pifont}
\newcommand{\cmark}{\ding{51}}
\newcommand{\xmark}{\ding{55}}

\usepackage{adjustbox}
\newcommand{\name}{0}
\newcommand{\h}{0}
\newcommand{\w}{0.15}

\newlength \g

\usepackage{enumitem}

\usepackage{graphicx}
\usepackage{amsmath}
\usepackage{amssymb}
\usepackage{booktabs}
\usepackage{balance}

\newcolumntype{L}[1]{>{\raggedright\let\newline\\\arraybackslash\hspace{0pt}}m{#1}}
\newcolumntype{C}[1]{>{\centering\let\newline\\\arraybackslash\hspace{0pt}}m{#1}}
\newcolumntype{R}[1]{>{\raggedleft\let\newline\\\arraybackslash\hspace{0pt}}m{#1}}

\usepackage[pagebackref,breaklinks,colorlinks]{hyperref}

\usepackage[capitalize]{cleveref}
\crefname{section}{Sec.}{Secs.}
\Crefname{section}{Section}{Sections}
\Crefname{table}{Table}{Tables}
\crefname{table}{Tab.}{Tabs.}

\def\cvprPaperID{8197} 
\def\confName{CVPR}
\def\confYear{2023}

\usepackage[accsupp]{axessibility}  
\begin{document}

\title{CiaoSR: Continuous Implicit Attention-in-Attention Network for\\ Arbitrary-Scale Image Super-Resolution}

\author{Jiezhang Cao$^{1}$ ~~~Qin Wang$^{1}$ ~~~Yongqin Xian$^{1}$\thanks{Currently with Google. This work was done at ETH Z\"urich.} ~~~Yawei Li$^{1}$ ~~~Bingbing Ni$^{2}$ ~~~Zhiming Pi$^{2}$\\ Kai Zhang$^{1}$\textsuperscript{\textdagger} ~~~Yulun Zhang$^{1}$\thanks{Corresponding Authors: Kai Zhang, cskaizhang@gmail.com; Yulun Zhang, yulun100@gmail.com} ~~~Radu Timofte$^{1,3}$ ~~~Luc Van Gool$^{1,4}$\\%
{$^{1}$ETH Z\"urich ~~~$^{2}$Huawei Inc. ~~~$^{3}$University of Wurzburg ~~~$^{4}$KU Leuven}\\
{\small \textbf{\url{https://github.com/caojiezhang/CiaoSR}}}
}
\maketitle

\begin{abstract}
Learning continuous image representations is recently gaining popularity for image super-resolution (SR) because of its ability to reconstruct high-resolution images with arbitrary scales from low-resolution inputs.
Existing methods mostly ensemble nearby features to predict the new pixel at any queried coordinate in the SR image. 
Such a local ensemble suffers from some limitations: i) it has no learnable parameters and it neglects the similarity of the visual features; ii) it has a limited receptive field and cannot ensemble relevant features in a large field which are important in an image.
To address these issues, this paper proposes a \textbf{c}ontinuous \textbf{i}mplicit \textbf{a}ttention-in-attenti\textbf{o}n network, called \textbf{CiaoSR}. We explicitly design an implicit attention network to learn the ensemble weights for the nearby local features. Furthermore, we embed a scale-aware attention in this implicit attention network to exploit additional non-local information.
Extensive experiments on benchmark datasets demonstrate CiaoSR significantly outperforms the existing single image SR methods with the same backbone.
In addition, CiaoSR also achieves the state-of-the-art performance on the arbitrary-scale SR task. The effectiveness of the method is also demonstrated on the real-world SR setting. More importantly, CiaoSR can be flexibly integrated into any backbone to improve the SR performance.
\end{abstract}

\vspace{-5mm}
\section{Introduction}
\label{sec:intro}
\vspace{-2mm}
Single image super-resolution (SISR), which aims to reconstruct a high-resolution (HR) image from a low-resolution (LR) one, has been widely employed in many practical applications \cite{gunturk2004super,zou2011very,shi2013cardiac}.
However, deep neural networks (DNN)-based SISR methods are facing some limitations in some real-world scenarios with arbitrary scales. For example, camera users may want to enhance the digital zoom quality by super-resolving a photo or a video to continuous arbitrary scales. 
Most existing DNN-based SISR methods \cite{lim2017edsr,zhang2018rcan,liang2021swinir} need to train a series of models for all different scales separately. However, 
it can be impractical to store all these models on the device due to limited storage and computing power.
Alternatively, arbitrary-scale image SR methods \cite{hu2019metasr,chen2021liif,lee2022lte} aim to train a single network for all scales in a continuous manner.

\begin{figure}[t]
  \centering
   \includegraphics[width=1\linewidth]{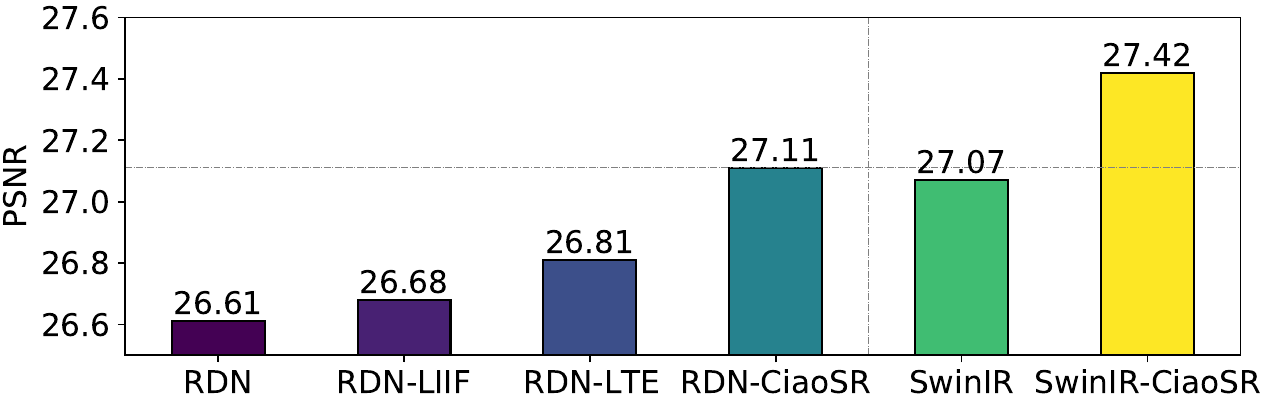}
   \vspace{-6mm}
   \caption{Comparison of different backbones and implicit models. Our proposed implicit neural network on RDN~\cite{zhang2018rdn} has better performance than SwinIR~\cite{liang2021swinir}. Check Section~\ref{sec:exp} for details.}
   \vspace{-0.6cm}
   \label{fig:comp_performance}
\end{figure}

\begin{figure*}[t]
   \centering
   \includegraphics[width=0.86\linewidth]{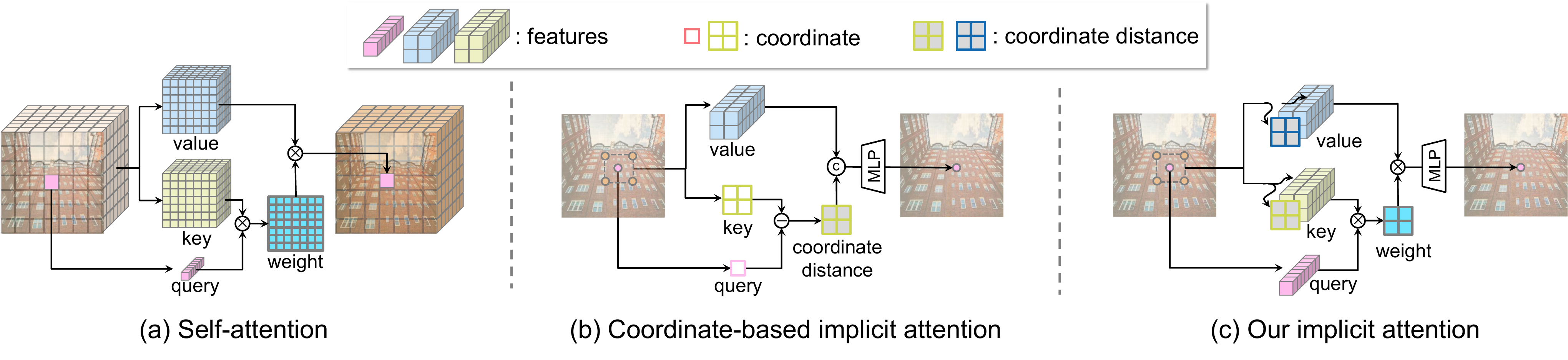}
   \vspace{-4mm}
   \caption{Comparisons of different attention mechanisms. (a) Self-attention can predict pixel features on the grid, but it cannot be directly used in arbitrary-scale SR without considering coordinates.  (b) Most existing methods can be treated as coordinate-based implicit attention since they calculate the distance between a key and query coordinate, and then use a function $g$ to aggregate with the value features. However, these methods ignore the distance between the features. (c) Our implicit attention not only considers the coordinate distance, but also the distance among features with visual information. 
   }
   \label{fig:comp_attn}
  \vspace{-5mm}
\end{figure*}

Most existing SISR methods \cite{lim2017edsr,zhang2018rcan,liang2021swinir} consist of a DNN and an upsampling module (\eg pixel shuffling \cite{shi2016pixelshuffle}) at a discrete scale. 
While substantial progresses have been made in the DNN backbones for SR, there is little attempt to study the upsampling module. 
A natural question to ask is: \emph{Does the pixel shuffling hinder the potential of SR models?}
One limitation of the pixel shuffling module is that it cannot synthesize SR images at large unseen and continuous scales.
To tackle this, one can treat synthesizing different-scale SR images as a multi-task learning problem, and train a specific upsampling module for each scale \cite{lim2017edsr}. 
However, these tasks are dependent and highly inter-related.
Neglecting the correlation of different-scale SR tasks may lead to discrete representations and limited performance.
Under a certain capacity of a network, training a model on multi-tasks may sacrifice the performance or have the comparable performance on each task.
These above disadvantages limit its applicability and flexibility in the real-world scenarios.

To address these, most existing arbitrary-scale SR methods \cite{chen2021liif,hu2019metasr,lee2022lte} replace the upsampling method with an implicit neural function and boost the performance.
These methods predict an RGB value at the query point in an image by ensembling features within a local region.
However, the local ensemble methods have limitations in the ensemble weights and insufficient information (\eg non-local information).
The ensemble weights are often calculated by the area of the rectangle between the query point and each nearest point, which is equivalent to the bilinear interpolation.
Thus, those methods cannot adaptively ensemble local features since there is no trainable parameter.
These weights are only related to the coordinates of the local features, but independent of the local features.
Ignoring both the coordinates and the local features lose visual information and result in blurry artifacts.
\emph{It is important and necessary to design a new implicit network to predict the weights and exploit more information in the local ensemble.}

In this paper, we propose a novel implicit attention model to enhance arbitrary-scale SR performance. Specifically, we use our attention to predict the ensemble weights by considering both the similarity and coordinate distance of local features, as shown in Figure \ref{fig:comp_attn}. Based on such learnable weights, the implicit model can adaptively aggregate local features according to different inputs.
To enrich more information, we introduce an attention in our implicit attention, which helps discover more features in a larger receptive field.

Our contributions are summarized as follows:
\vspace{-5pt}
\begin{itemize}[leftmargin=3mm]
    \setlength{\itemsep}{0pt}
    \item We propose a novel continuous implicit attention-in-attention network for arbitrary-scale image SR, called CiaoSR. Different from most existing local ensemble methods, our method explicitly learns the ensemble weights and exploits scale-aware non-local information.
    \item Our CiaoSR can be flexibly integrated into any backbone, allowing the network to super-resolve an image at arbitrary scales and improve the SR performance in Figure \ref{fig:comp_performance}. 
    \item Extensive experiments demonstrate CiaoSR achieves the state-of-the-art performance in both SISR and arbitrary-scale SR tasks. 
    Besides, our CiaoSR has good generalization on both in-scale and out-of-scale distributions.
    Last, we extend our method to real-world SR settings to synthesize arbitrary-scale images.
\end{itemize}

\begin{figure*}
  \begin{center}
  \includegraphics[width=0.93\linewidth]{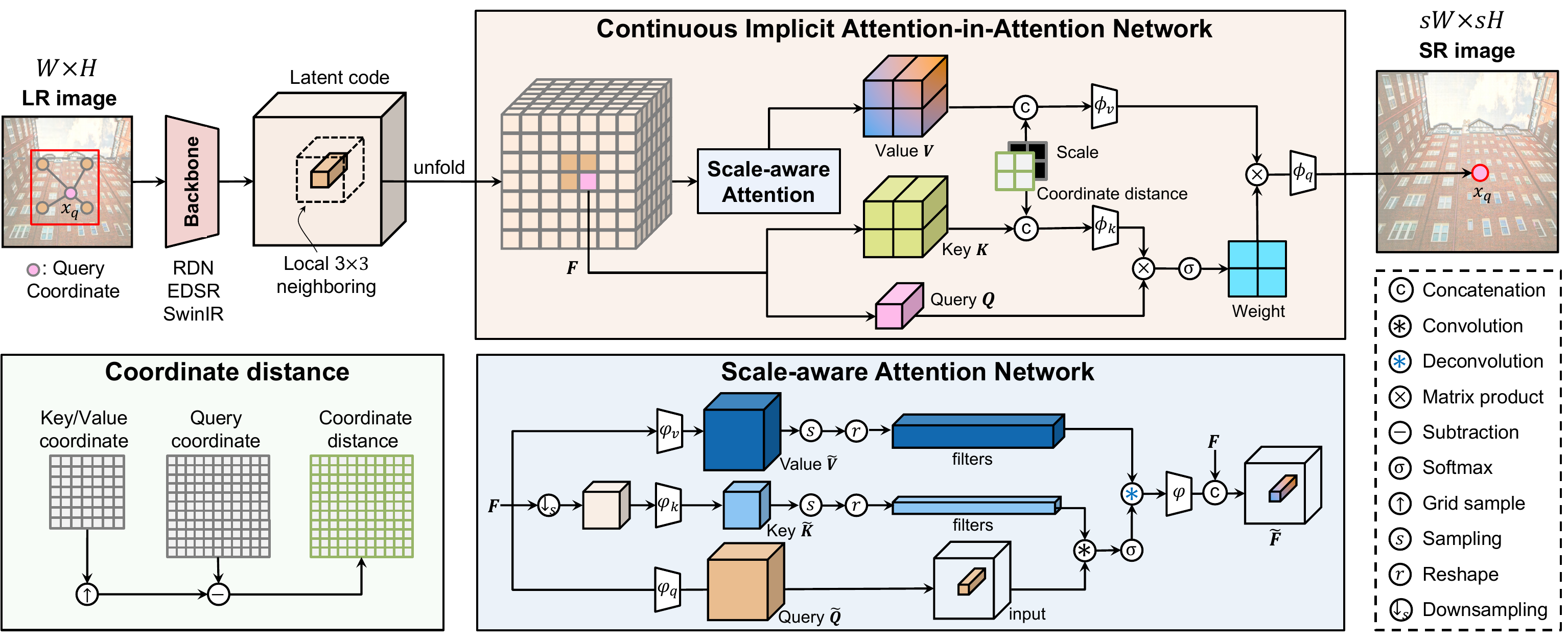}
  \end{center}
  \vspace{-6mm}
  \caption{The architecture of our continuous implicit attention-in-attention network. 
  Given an LR image, the encoder extracts features as latent codes. For a query point, we have a query feature and key features close to the query point, and the scale-aware non-local attention module extracts non-local features as value. Last, we use the triple of query, key and value to predict the RGB value at the query point.}
  \label{fig:arch}
  \vspace{-4mm}
\end{figure*}

\section{Related Work}
\label{sec:related_work}

\vspace{-2mm}
\noindent\textbf{Single image super-resolution (SISR).}
SISR aims to synthesize high-resolution (HR) images from low-resolution (LR) images.
Compared with DNN-based SR methods \cite{gu2012fast, timofte2013anchored, timofte2014a, michaeli2013nonparametric, he2010darkchannel, cao2022datsr, cao2022davsr, cao2021vsrt}, methods in recent years build on deep convolutional neural network (CNN) to improve the performance, such as SRCNN~\cite{dong2014srcnn}, SRResNet~\cite{ledig2017srresnet}, EDSR~\cite{lim2017edsr}, RDN~\cite{zhang2018rdn} and RCAN~\cite{zhang2018rcan}.
To further improve SR performance, some methods design CNN with residual block~\cite{kim2016vdsr, cavigelli2017cas, zhang2021DPIR}, dense block~\cite{wang2018esrgan, zhang2018rdn, zhang2020RDNIR} and others ~\cite{chen2016TNRD, lai2017LapSRN, zhang2018srmd, wang2019learning, wang2021unsupervised, wang2021learning, liang2021fkp, liang21hcflow, liang21manet, zhang2018ffdnet, tai2017memnet, isobe2020video, wei2021unsupervised, guo2020closed, cheng2021mfagan, deng2021deep, zhang2019RNAN, peng2019dsnet, jia2019focnet, fu2019jpeg, kim2019pseudo, fu2021model, zhou2020IGNN}.
In addition, some SR methods are built based on attention mechanism~\cite{vaswani2017transformer}, such as channel attention~\cite{zhang2018rcan, dai2019SAN, niu2020HAN}, self-attention (IPT~\cite{chen2021IPT} and SwinIR~\cite{liang2021swinir}, HAT~\cite{chen2022hat}), non-local attention~\cite{liu2018NLRN, mei2021NLSA}. 
However, most methods focus on one specific scale, which limits the applicability and flexibility in arbitrary-scale.

\vspace{0.5mm}
\noindent\textbf{Arbitrary-scale super-resolution.}
To tackle this problem, 
very recently, the more practical setup of arbitrary-scale SR is considered, which aims to super-resolute images with arbitrary scales by a single model.
MetaSR~\cite{hu2019metasr} makes the first attempt to propose an arbitrary-scale meta-upscale module.
To improve the performance, many arbitrary-scale methods \cite{wang2021arbsr,son2021srwarp} are proposed.
With the help of implicit neural representation \cite{sitzmann2020SIREN,chen2019learning,michalkiewicz2019implicit,atzmon2020sal,gropp2020implicit,sitzmann2019scene,jiang2020local,peng2020convolutional,chabra2020deep,niemeyer2020differentiable,oechsle2019texture,mildenhall2021nerf}, LIIF~\cite{chen2021liif} predicts the RGB value at an arbitrary query coordinate by taking an image coordinate and features of backbone around the coordinate.
The features are extracted by single image super-resolution methods, \eg EDSR~\cite{lim2017edsr}, RDN~\cite{zhang2018rdn} and SwinIR~\cite{liang2021swinir}.
To improve the performance, existing methods~\cite{lee2022lte,xu2021ultrasr} propose to integrate more features in SR models.
For example, LTE~\cite{lee2022lte} proposes a local texture estimator by characterizing image textures in the Fourier space.
UltraSR~\cite{xu2021ultrasr} integrates spatial coordinates and periodic encoding in the implicit network. 
These methods use the bilinear interpolation to ensemble nearby features.
However, such an ensemble way has no learnable parameters.
Recently, ITSRN \cite{yang202itsrn} learns the weights by taking the coordinate distance and scale token into a mapping.
Most methods learn the ensemble without the feature similarity. 


\section{Preliminary and Motivation}\label{sec:pre}
\vspace{-1mm}
Let $\bI$ be a continuous image, and $\bx$ be a 2D coordinate of a pixel in the image $\bI$.
Formally, given a 2D coordinate $\bx$ in the continuous image domain and latent code $\bZ$ extracted by deep neural networks, the RGB value can be predicted by an implicit image function which can be defined as follows,
\begin{equation}
    {\bI}(\bx) = f (\bZ, \bx),  \label{eq:implicit}
    \vspace{-1mm}
\end{equation}
where the implicit image function can be parameterized by a multilayer perceptron (MLP).
Note that this implicit image function is shared by all images.
For recent SISR methods \cite{zhang2018rcan,liang2021swinir}, the implicit image function $f$ can be 
implemented as a PixelShuffle~\cite{shi2016pixelshuffle} with convolutions with a specific scale.
However, these methods are independent on the coordinates, leading to an issue that they only adapt to the specific scale and are inflexible to synthesize arbitrary-scale images.

To predict the RGB value $\bI_q$ at an arbitrary query coordinate $\bx_q$, 
most existing methods \cite{chen2021liif,lee2022lte,yang202itsrn} propose to compute the RGB value at coordinate $\bx_q$ by directly ensembling its neighborhood information, 
\begin{align} \label{eq:ensemble}
    \bI_q := \bI(\bx_q) = \sum\nolimits_{(i, j)\in\mI} ~w_{i,j} \cdot f(\bZ_{i,j}^*, \bx_q-\bx_{i,j}^*),
\end{align}
where $\mI$ is the local region centered at the query coordinate $\bx_q$, \eg $\mI$ can be top-left, top-right, bottom-left, bottom-right coordinates, 
and $w_i$ is the weight of the neighboring pixel $\bx^*_{i,j}$, which is calculated \wrt the area of the rectangle between $\bx_q$ and $\bx_{i,j}^*$, as shown in Figure \ref{fig:comp_ensemble}. 
However, the performance improvement is limited because the ensemble weight $w_{i,j}$ is purely based on the coordinates. The visual similarities are completely ignored in the weight calculation.
Besides, these methods only consider the nearest latent codes, leading to a limited receptive field.
In this paper, our goal is to \textit{learn} the weights adaptively by leveraging both visual information and coordinate information.

\vspace{-2mm}
\section{Proposed Method}
\label{sec:method}
\vspace{-1mm}
The above local ensemble Eqn. \eqref{eq:ensemble} has a similar form to the attention mechanism.
Specifically, the weights $w_i$ in Eqn. \eqref{eq:ensemble} can be modeled as an attention map and the latter term $f$ can be the value in the attention mechanism.
Such an attention map is related to the similarities of the latent code and coordinate, and it can be calculated using both query and key which can integrate the latent code and coordinate information.
In this sense, attention models can be used to mitigate the drawbacks of previous methods~(\ie calculated purely based on coordinates without considering any visual information) by learning the soft weights from both visual and coordinate information.
However, the use of attention in the implicit function is non-trivial because standard self-attention  \cite{vaswani2017transformer} and neighborhood attention \cite{hassani2022neighborhood} mechanisms are based on visual features, and not conditioned on the coordinate information.  
To exploit the continuous representation learning from both the coordinate information and visual features, we propose a new attention for arbitrary-scale SR.

\vspace{1mm}
\noindent\textbf{Continuous implicit attention-in-attention.}
The architecture of the proposed continuous implicit attention-in-attention for arbitrary-scale SR (called \textbf{CiaoSR}) is shown in Figure \ref{fig:arch}.
It is called attention-in-attention because the value contains an additional embedded attention that additionally captures non-local information of repetitive patterns.
Specifically, given an LR image,  our attention is provided with both the visual features and coordinate information to \textit{learn} good ensemble weights for the local neighboring latent codes for any query at any arbitrary-scale $s$.

Given a query coordinate $\bx_q$ in the upsampled image, the known LR pixels in the local region centered at $\bx_q$ are considered for the local ensemble. We define $\bx_k$ and $\bx_v$ as the key and value coordinates of the local neighbor.  
Let $\bQ, \bK$ and $\bV$ be the latent code of query, key and value on the corresponding coordinates, respectively.
Given pairs of coordinates and latent codes $(\bQ, \bx_q)$, $(\bK, \bx_k)$ and $(\bV, \bx_v)$, we define a new implicit attention function (called $\mbox{i-Attention}$) to predict RGB values at the given query $\bx_q$,
\begin{align}
    \bI_q = \mbox{i-Attention} (\bQ, \bK, \bV; \bx_q, \bx_k, \bx_v),
\end{align}
where the value feature $\bV$ can be embedded with another attention to aggregate more non-local information. 

As shown in Figure \ref{fig:comp_attn}, our implicit attention is different from existing attention mechanisms by considering the additional coordinate information.
In particular, the self-attention mechanism \cite{vaswani2017transformer} without considering coordinates cannot predict pixel features that are not on the grid. 
The coordinate-based implicit attention calculates the distance between a key $\bx_k$ and query coordinate $\bx_q$, which ignores the visual information and similarity of the features. 

\vspace{1mm}
\noindent\textbf{Implicit attention to learn weights for the local ensemble.}
Unlike previous methods~\cite{chen2021liif,lee2022lte} which directly use the normalized area as the ensemble weights, we propose to first use an attention to learn better weights for the ensemble of the local region. Our attention takes both the coordinate distance $\br$, the local features $\bF_l$, as well as non-local features $\bG$ into account for the learning of the weight. Formally, the attention can be written as
\begin{align}
    \bI_q = \phi_q \Big( \sum\nolimits_{(i, j)\in\mI} \underbrace{\sigma(\bQ^{\top}\bK_{i, j})}_{\text{ensemble weight}} \bV_{i,j} \Big), \label{eqn:Iq}
\end{align}
where $\phi_q$ is the query network, $\sigma$ is the Softmax function, and $\mI$ is the local region centered at $\bx_q$. The query, key, and value for this local ensemble attention are defined as
\begin{align}
    \left\{
    \begin{aligned}
    \bQ &= \bF^*, \\
    \bK_{i, j} &= \phi_k([\bF_{i, j}, (\br_{k})_{i,j}, \bs]), \\
    \bV_{i, j} &= \phi_v([\bF_{i, j}, \tilde{\bF}_{i, j}], (\br_{v})_{i,j}, \bs]),
    \end{aligned}
    \right.
\end{align}
where $\phi_k$ and $\phi_v$ are the key and value networks, respectively, $\bF^*$ are the features of nearest neighbors on the grid of the LR image \wrt the query coordinate $\bx_q$, $\bF_{i, j}$ is a local feature calculated by using the unfolding operator to the latent codes at the $(i, j)$-th position, which is equivalent to concatenating the latent codes in neighboring $3{\times}3$ region.
$\tilde{\bF}_{i, j}$ are non-local features, which will be introduced below in detail. $\bs = [s_h; s_w]$ is a two-dimensional scale, and $\br_k$ and $\br_v$ are the relative distance between the coordinates in the local region, \ie
\begin{align}
    \br_{k} = \bx_q - (\bx_k)_{i,j}, \quad \br_{v} = \bx_q - (\bx_v)_{i,j}.
\end{align}

\begin{figure}[t]
  \centering
   \includegraphics[width=0.89\linewidth]{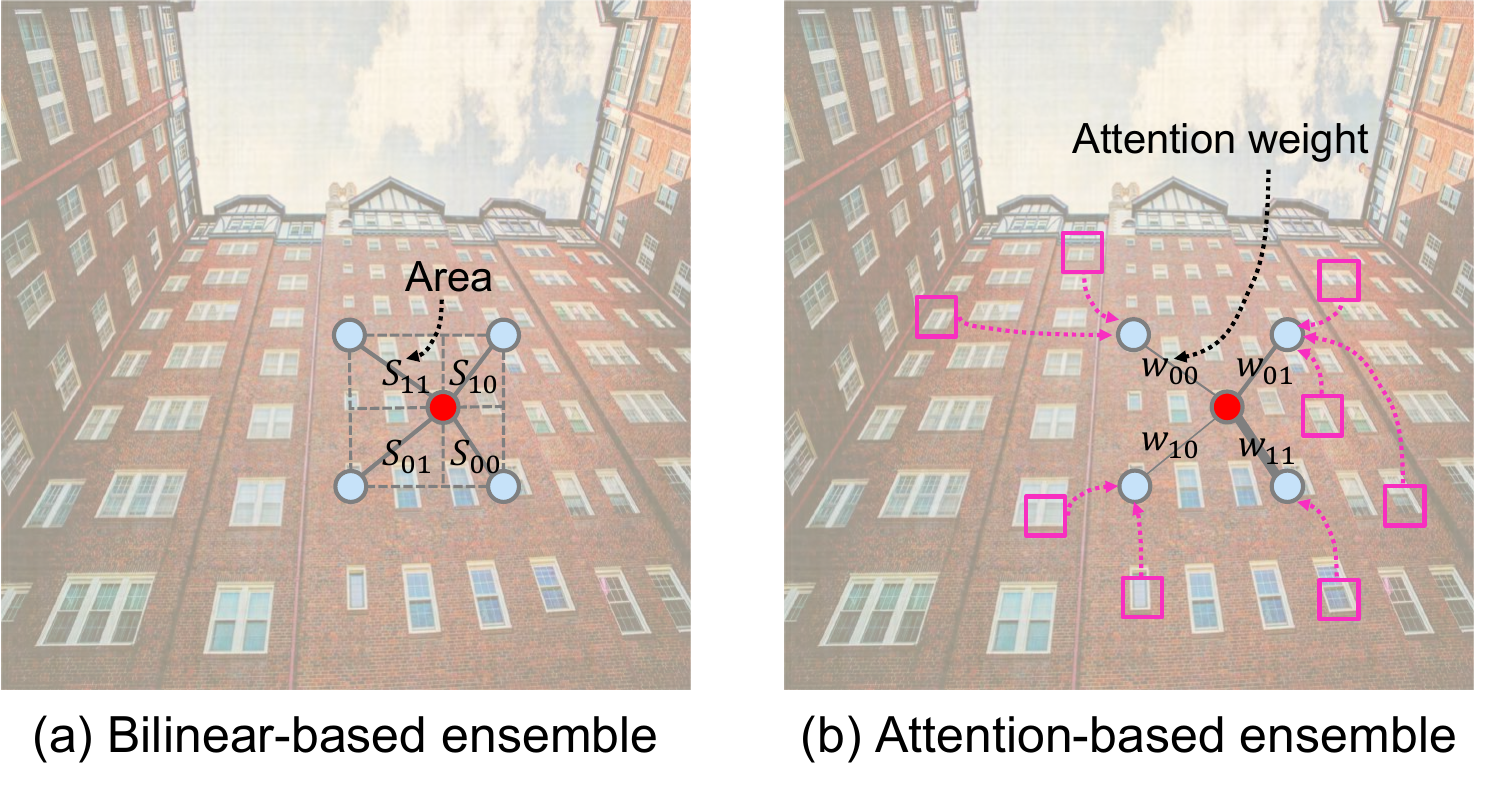}
   \vspace{-4mm}
   \caption{Comparison of different ensemble methods. (a) Most existing methods calculate the weights related to the area of the rectangle between the query coordinate and the nearest point. Such weights are equivalent to the weight of the bilinear interpolation. (b) Our attention-based ensemble calculates the attention map as the weights for feature aggregation. Besides, the ensemble method has a large receptive field for exploiting more information.}
   \vspace{-0.4cm}
   \label{fig:comp_ensemble}
\end{figure}

\noindent\textbf{Embedded scale-aware attention for non-local features.}
Given a query coordinate, the local ensemble attention  only considers the keys and values in the small local region, resulting in a limited receptive field. 
However, there may exist similar textures with different scales within a single image in other positions. For example, repetitive patterns in different scales~(\eg facades, windows, etc. in a building, as shown in Figure \ref{fig:comp_ensemble}) can exist in different locations within one image. 
To exploit the useful non-local information, we propose an embedded scale-aware non-local attention module, which aggregates the information from all coordinates on the grid of the LR image, motivated by \cite{mei2020image}. 
Specifically, we obtain query $\tilde{\bQ}$ and value features $\tilde{\bV}$ from the latent code $\bF$ of the LR image, and downsample $\bF$ to a scale $s'$ as a key feature $\tilde{\bK}$.
We first calculate the pixel-wise similarity between the query located at $(i,j)$ and the key on other coordinates, and then aggregate the weighted information from all of them.
Lastly, we use a convolutional layer $\varphi$ for downsampling with the scale $s'$ normalized similarity matrix to aggregate the features.
Formally, the non-local feature $\tilde{\bF}$ located at $(i,j)$ can be calculated by
\begin{align}
    \!\!\tilde{\bF}_{i, j} {=} \varphi \left( \sum\limits_{u,v} \frac{ \exp \left(\tilde{\bQ}_{i,j}^{\top} \tilde{\bK}_{u, v}\right) }{\sum\nolimits_{u', v'} \exp \left(\tilde{\bQ}_{i,j}^{\top} \tilde{\bK}_{u', v'} \right)} \tilde{\bV}_{s'u, s'v}^{s'p{\times}s'p} \right),
\end{align}
where $\tilde{\bV}_{s'u, s'v}^{s'p{\times}s'p}$ is the value feature patch of the size $s'p{\times}s'p$ located at $(s'u, s'v)$, $\tilde{\bQ}, \tilde{\bK}$ and $\tilde{\bV}$ are the query, key and value's non-local features, which can be calculated by 
\begin{align}
    \left\{
    \begin{aligned}
    \tilde{\bQ} &= \varphi_q(\bF), \\
    \tilde{\bK} &= \varphi_k(\bF \downarrow_{s'}), \\
    \tilde{\bV} &= \varphi_v(\bF),
    \end{aligned}
    \right.
\end{align}
where $\downarrow_{s'}$ is the downsampling operation with a given scale $s'$ which can be $\{2, 3, 4\}$, $\varphi_q$, $\varphi_k$ and $\varphi_v$ are query, key and value networks.
Different from \cite{mei2020image}, our method can integrate multi-scale features.
In the experiment, we can search for relevant features in a small window (\eg $256{\times}256$), instead of the entire image.
In this way, our method can guarantee a large receptive field, and reduce computational cost.

\begin{table*}[t]
\caption{Quantitative comparison with state-of-the-art methods for \underline{\textbf{arbitrary-scale SR}} on the DIV2K validation set (PSNR (dB)). \textbf{Bold} indicates the best performance. 
$\dag$ indicates the implementation of \cite{lee2022lte}. 
}
\vspace*{-1pt}
\label{tab:comp_DIV2K}
\centering
\resizebox{1\textwidth}{!}{
\begin{tabular}{l|l||ccc|ccccc}
\hline\toprule
\rowcolor{mygray}
& & \multicolumn{3}{c|}{$\qquad$In-scale$\qquad$} & \multicolumn{5}{c}{Out-of-scale} \\
\cmidrule{3-10}
\rowcolor{mygray}
\multirow{-2}{*}{Backbones} & \multirow{-2}{*}{Methods} &~~~~~$\times2$~~~~~ & ~~~~~$\times3$~~~~~ & ~~~~~$\times4$~~~~~ & ~~~$\times6$~~~ & ~~~$\times12$~~~ & ~~~$\times18$~~~ & ~~~$\times24$~~~ & ~~~$\times30$~~~ 
\\
\hline\hline
\multirow{1}{*}{-} &
{Bicubic} & 31.01 & 28.22 & 26.66 & 24.82  & 22.27 & 21.00 & 20.19 & 19.59 \\
\hline
\multirow{6}{*}{EDSR \cite{lim2017edsr}} &
EDSR-baseline \cite{lim2017edsr} & 34.55 & 30.90 & 28.94 & - & - & - & - & - \\
& EDSR-baseline-MetaSR \cite{hu2019metasr}~~ & 34.64 & 30.93 & 28.92 & 26.61 & 23.55 & 22.03 & 21.06 & 20.37 \\
& EDSR-baseline-LIIF \cite{chen2021liif} & {34.67} & {30.96} & {29.00} & {26.75} & {23.71} & {22.17} & {21.18} & {20.48} \\
& EDSR-baseline-ITSRN$^{\dag}$ \cite{yang202itsrn} &
34.71 & 30.95 & 29.03 & 26.77 & 23.71 & 22.17 & 21.18 & 20.49 \\
& EDSR-baseline-LTE \cite{lee2022lte} & {34.72} & {31.02} & {29.04} & {26.81} & {23.78} & {22.23} & {21.24} & {20.53} \\
& \bf{EDSR-baseline-CiaoSR (ours)~~~~~} & \bf{34.91} & \bf{31.15}  &  \bf{29.23} & \bf{26.95}  & \bf{23.88} & \bf{22.32} & \bf{21.32} & \bf{20.59}  \\
\hline
\multirow{6}{*}{RDN \cite{zhang2018rdn}} &
RDN-baseline \cite{zhang2018rdn} & 34.94 & 31.22 & 29.19 & - & - & - & - & - \\
& RDN-MetaSR \cite{hu2019metasr} & {35.00} & {31.27} & 29.25 & 26.88 & 23.73 & 22.18 & 21.17 & 20.47 \\
& RDN-LIIF \cite{chen2021liif} & 34.99 & 31.26 & {29.27} & {26.99} & {23.89} & {22.34} & {21.31} & {20.59} \\
& RDN-ITSRN$^{\dag}$ \cite{yang202itsrn} &
35.09 & 31.36 & 29.38 & 27.06 & 23.93 & 22.36 & 21.32 & 20.61 \\
& RDN-LTE \cite{lee2022lte} & {35.04} & {31.32} & {29.33} & {27.04} & {23.95} & {22.40} & {21.36} & {20.64} \\
& \bf{RDN-CiaoSR (ours)} &
\bf{35.15} & \bf{31.42} & \bf{29.45} & \bf{27.16} &
\bf{24.06}  & \bf{22.48}  & \bf{21.43}  & \bf{20.70}  \\
\hline
\multirow{6}{*}{SwinIR \cite{liang2021swinir}} &
SwinIR-baseline \cite{liang2021swinir} & 34.94 & 31.22 &  29.19 &  - & - & - & - & - \\
& SwinIR-MetaSR$^{\dag}$ \cite{hu2019metasr} & 35.15 & 31.40 & 29.33 & 26.94 & 23.80 & 22.26 & 21.26 & 20.54 \\
& SwinIR-LIIF$^{\dag}$ \cite{chen2021liif} & {35.17} & {31.46} & {29.46} & {27.15} & {24.02} & {22.43} & {21.40} & {20.67} \\
& SwinIR-ITSRN$^{\dag}$ \cite{yang202itsrn} & 35.19 & 31.42 
& 29.48 & 27.13 & 23.83 & 22.31 & 21.31 & 20.55 \\
& SwinIR-LTE \cite{lee2022lte} & {35.24} & {31.50} & {29.51} & {27.20} & {24.09} & {22.50} & {21.47} & {20.73} \\
& \bf{SwinIR-CiaoSR (ours)} & \bf{35.29}  & \bf{31.55} & \bf{29.59} & \bf{27.28} & \bf{24.15} & \bf{22.54} & \bf{21.51} & \bf{20.74} \\
\bottomrule
\end{tabular}
}
\vspace{-5mm}
\end{table*}

\newpage
\section{Experiments}
\label{sec:exp}

\vspace{-5pt}
\noindent\textbf{Datasets.}
For the arbitrary-scale SR task, we follow \cite{chen2021liif,lee2022lte} and use DIV2K \cite{DIV2K} as the training set, which consists of 800 images in 2K resolution.
In testing, we evaluate the models on the DIV2K validation set and a wide range of standard benchmark datasets, including Set5 \cite{Set5}, Set14 \cite{Set14}, B100 \cite{BSD100}, Urban100 \cite{Urban100}, and Manga109 \cite{Manga109}. 

\vspace{1pt}
\noindent\textbf{Evaluation metrics.}
We use PSNR to evaluate the quality of the synthesized SR images.
Following \cite{chen2021liif,lee2022lte}, the PSNR value is calculated on  all RGBs channels for the DIV2K validation set, and additionally also calculated on the Y channel (\ie luminance) of the transformed YCbCr space for other benchmark test sets.
In addition, other evaluation metrics (such as SSIM \cite{wang2004ssim} and LPIPS \cite{zhang2018lpips}) are provided to evaluate the image quality.

\vspace{1pt}
\noindent\textbf{Implementation details.}
We follow the same experimental setup as prior works \cite{chen2021liif,lee2022lte}.
To synthesize paired training data, we first crop the $48s{\times}48s$ patches as ground-truth (GT) images, and use the Bicubic downsampling in PyTorch to have LR images with the patch size of $48{\times}48$, where $s$ is a scale factor sampled in the uniform distribution $\mU(1, 4)$.
We sample $48^2$ pixels from both GT images and the corresponding coordinates.
We use the existing SR models (\eg EDSR \cite{lim2017edsr}, RDN \cite{zhang2018rdn} and SwinIR \cite{liang2021swinir}) as backbones by removing their upsampling modules.
The detailed architecture of our implicit network is provided in the supplementary material.
For the training, we use Adam \cite{kingma2014adam} as the optimizer, and use L1 loss to train all models for 1000 epochs with the batch size of 16 per GPU.
We set the learning rate as $1e-4$ at the beginning and decay the learning rate by a factor of 0.5 every 200 epochs.
These experimental settings are identical to \cite{chen2021liif,hu2019metasr,lee2022lte}.

\begin{table*}[t!]
\centering
\vspace{-1pt}
\caption{Quantitative comparison with state-of-the-art methods for \underline{\textbf{in-scale SR}} on benchmark datasets (PSNR (dB)). \textbf{Bold} indicates the best performance. $\dag$ indicates our implementation.
}
\label{tab:comp_benchmark_in}
\resizebox{1\textwidth}{!}{
\begin{tabular}{l||ccc|ccc|ccc|ccc|ccc}
\hline
\toprule
\rowcolor{mygray}
& \multicolumn{3}{c|}{Set5 \cite{Set5}} 
& \multicolumn{3}{c|}{Set14 \cite{Set14}} 
& \multicolumn{3}{c|}{B100 \cite{BSD100}} 
& \multicolumn{3}{c|}{Urban100 \cite{Urban100}}
& \multicolumn{3}{c}{Manga109 \cite{Manga109}}
\\ 
\cmidrule{2-16}
    \rowcolor{mygray}
    \multirow{-2}{*}{Methods} 
    & $\times 2$ & $\times 3$ & $\times 4$ 
    & $\times 2$ & $\times 3$ & $\times 4$ 
    & $\times 2$ & $\times 3$ & $\times 4$ 
    & $\times 2$ & $\times 3$ & $\times 4$
    & $\times 2$ & $\times 3$ & $\times 4$ \\ 
    \hline\hline
    RDN \cite{zhang2018rdn}
    & 38.24
    & 34.71
    & 32.47
    & 34.01
    & 30.57
    & 28.81
    & 32.34
    & 29.26
    & 27.72
    & 32.89
    & 28.80
    & 26.61
    & 39.18
    & 34.13
    & 31.00
    \\  
    RDN-MetaSR \cite{hu2019metasr}
    & 38.22
    & 34.63
    & 32.38
    & 33.98
    & 30.54
    & 28.78  
    & 32.33
    & 29.26
    & 27.71
    & 32.92
    & 28.82
    & 26.55
    & -
    & -
    & -
    \\
    RDN-LIIF \cite{chen2021liif}
    & 38.17
    & 34.68
    & 32.50
    & 33.97
    & 30.53
    & 28.80
    & 32.32
    & 29.26
    & 27.74
    & 32.87
    & 28.82
    & 26.68
    & 39.26
    & 34.21
    & 31.20
    \\
    RDN-ITSRN$^{\dag}$ \cite{yang202itsrn}
    & 38.23
    & 34.76
    & 32.55
    & 34.19
    & 30.59
    & 28.88 
    & 32.38
    & 29.32
    & 27.79
    & 33.07
    & 28.96
    & 26.77
    & 39.34
    & 34.39
    & 31.37
    \\
    RDN-LTE \cite{lee2022lte}
    & 38.23
    & 34.72
    & 32.61
    & 34.09 
    & 30.58 
    & 28.88  
    & 32.36 
    & 29.30 
    & 27.77
    & 33.04 
    & 28.97 
    & 26.81 
    & 39.28
    & 34.32
    & 31.30
    \\
    \bf{RDN-CiaoSR (ours)}
    & \bf{38.29}
    & \bf{34.85}
    & \bf{32.66}
    & \bf{34.22}
    & \bf{30.65}
    & \bf{28.93} 
    & \bf{32.41}
    & \bf{29.34}
    & \bf{27.83}
    & \bf{33.30}
    & \bf{29.17}
    & \bf{27.11}
    & \bf{39.51}
    & \bf{34.57}
    & \bf{31.57}
    \\
    \cline{1-16}  
    SwinIR \cite{liang2021swinir}
    & 38.35
    & 34.89
    & 32.72
    & 34.14 
    & 30.77 
    & 28.94 
    & 32.44 
    & 29.37 
    & 27.83
    & 33.40 
    & 29.29 
    & 27.07 
    & 39.60
    & 34.74
    & 31.67
    \\ 
    SwinIR-MetaSR \cite{hu2019metasr}
    & 38.26 
    & 34.77 
    & 32.47
    & 34.14 
    & 30.66 
    & 28.85 
    & 32.39 
    & 29.31 
    & 27.75
    & 33.29 
    & 29.12 
    & 26.76
    & 39.46
    & 34.62
    & 31.37
    \\
    SwinIR-LIIF \cite{chen2021liif}
    & 38.28 
    & 34.87 
    & 32.73 
    & 34.14 
    & 30.75 
    & 28.98 
    & 32.39 
    & 29.34 
    & 27.84 
    & 33.36 
    & 29.33 
    & 27.15
    & 39.57
    & 34.68
    & 31.71
    \\
    SwinIR-ITSRN$^{\dag}$ \cite{yang202itsrn}
    & 38.22
    & 34.75
    & 32.63
    & 34.26
    & 30.75
    & 28.97 
    & 32.42
    & 29.38
    & 27.85 
    & 33.46
    & 29.34
    & 27.12
    & 39.60
    & 34.75
    & 31.74
    \\
    SwinIR-LTE \cite{lee2022lte}
    & 38.33 
    & 34.89
    & 32.81 
    & 34.25 
    & 30.80 
    & 29.06 
    & 32.44 
    & 29.39 
    & 27.86
    & 33.50 
    & 29.41 
    & 27.24 
    & 39.63
    & 34.79
    & 31.79
    \\
    \bf{SwinIR-CiaoSR (ours)}
    & \bf{38.38}
    & \bf{34.91} 
    & \bf{32.84}
    & \bf{34.33}
    & \bf{30.82}
    & \bf{29.08} 
    & \bf{32.47}
    & \bf{29.42}
    & \bf{27.90}
    & \bf{33.65}
    & \bf{29.52}
    & \bf{27.42}
    & \bf{39.67}
    & \bf{34.84}
    & \bf{31.91}
    \\
    \bottomrule
\end{tabular}
    }
\end{table*}

\begin{table*}[t!]
\centering
\vspace{-3pt}
\caption{Quantitative comparison with state-of-the-art methods for \underline{\textbf{out-of-scale SR}} on benchmark datasets (PSNR (dB)). \textbf{Bold} indicates the best performance.  $\dag$ indicates our implementation.
}
\label{tab:comp_benchmark_out}
\resizebox{1\textwidth}{!}{
\begin{tabular}{l||ccc|ccc|ccc|ccc|ccc}
\hline\toprule
\rowcolor{mygray}
& \multicolumn{3}{c|}{Set5 \cite{Set5}} 
& \multicolumn{3}{c|}{Set14 \cite{Set14}} 
& \multicolumn{3}{c|}{B100 \cite{BSD100}} 
& \multicolumn{3}{c|}{Urban100 \cite{Urban100}}
& \multicolumn{3}{c}{Manga109 \cite{Manga109}}
\\ 
\cmidrule{2-16}
    \rowcolor{mygray}
    \multirow{-2}{*}{Methods} 
    &$\times 6$ & $\times 8$ & $\times 12$ 
    &$\times 6$ & $\times 8$ & $\times 12$
    &$\times 6$ & $\times 8$ & $\times 12$
    &$\times 6$ & $\times 8$ & $\times 12$
    &$\times 6$ & $\times 8$ & $\times 12$\\ \hline\hline
    RDN-MetaSR \cite{hu2019metasr}
    & 29.04
    & 29.96
    & -
    & 26.51
    & 24.97
    & -
    & 25.90
    & 24.83
    & -
    & 23.99
    & 22.59
    & -
    & -
    & -
    & -
    \\
    RDN-LIIF \cite{chen2021liif}
    & 29.15
    & 27.14
    & 24.86
    & 26.64
    & 25.15
    & 23.24
    & 25.98
    & 24.91
    & 23.57
    & 24.20
    & 22.79
    & 21.15
    & 27.33
    & 25.04
    & 22.36
    \\
    RDN-ITSRN$^{\dag}$ \cite{yang202itsrn}
    & 29.32
    & 27.25
    & 24.86
    & 26.68
    & 25.17
    & 23.28
    & 26.01
    & 24.93
    & 23.58
    & 24.23
    & 22.81
    & 21.16
    & 27.45
    & 25.04
    & 23.35
    \\
    RDN-LTE \cite{lee2022lte}
    & 29.32
    & 27.26
    & 24.79
    & 26.71 
    & 25.16
    & 23.31
    & 26.01 
    & 24.95
    & 23.60
    & 24.28 
    & 22.88
    & 21.22
    & 27.49
    & 25.12
    & 22.43
    \\
    \bf{RDN-CiaoSR (ours)}
    & \bf{29.46}
    & \bf{27.36} 
    & \bf{24.92}
    & \bf{26.79}
    & \bf{25.28}
    & \bf{23.37}
    & \bf{26.07}
    & \bf{25.00}
    & \bf{23.64}
    & \bf{24.58}
    & \bf{23.13}
    & \bf{21.42}
    & \bf{27.70}
    & \bf{25.40}
    & \bf{22.63}
    \\ \hline
    SwinIR-MetaSR \cite{hu2019metasr}
    & 29.09 
    & 27.02
    & 24.82
    & 26.58 
    & 25.09
    & 23.33
    & 25.94 
    & 24.87
    & 23.59
    & 24.16 
    & 22.75
    & 21.31
    & 27.29
    & 24.96
    & 22.35
    \\
    SwinIR-LIIF \cite{chen2021liif}
    & 29.46 
    & 27.36
    & -
    & 26.82 
    & 25.34
    & -
    & 26.07 
    & 25.01
    & -
    & 24.59 
    & 23.14
    & -
    & 27.69
    & 25.28
    & -
    \\
    SwinIR-ITSRN$^{\dag}$ \cite{yang202itsrn}
    & 29.31
    & 27.24
    & 24.79
    & 26.71
    & 25.32
    & 23.30
    & 26.05
    & 24.96
    & 23.57
    & 24.50
    & 23.06
    & 21.34
    & 27.72
    & 25.23
    & 22.47
    \\
    SwinIR-LTE \cite{lee2022lte}
    & 29.50 
    & 27.35
    & - 
    & 26.86 
    & \bf{25.42}
    & - 
    & 26.09 
    & 25.03
    & - 
    & 24.62 
    & 23.17
    & - 
    & 27.83
    & 25.42
    & - 
    \\
    \bf{SwinIR-CiaoSR (ours)}
    & \bf{29.62}
    & \bf{27.45}
    & \bf{24.96}
    & \bf{26.88}
    & \bf{25.42} 
    & \bf{23.38}
    & \bf{26.13}
    & \bf{25.07}
    & \bf{23.68}
    & \bf{24.84}
    & \bf{23.34}
    & \bf{21.60}
    & \bf{28.01}
    & \bf{25.61}
    & \bf{22.79}
    \\ \hline
\end{tabular}
    } 
\end{table*}

\subsection{Comparisons with State-of-the-Art}
\vspace{-2mm}
\paragraph{Quantitative results.}
In Tables \ref{tab:comp_DIV2K}-\ref{tab:comp_benchmark_out}, CiaoSR achieves the best performance with the highest PSNR across all datasets with all backbones concerned on both in-scale and out-of-scale distributions.
For the in-scales, in particular, our model has a remarkable PSNR gain of 0.3dB on Urban100 ($\times4$), compared with the previous SOTA method \cite{lee2022lte} under the same RDN backbone.
It is worth noting that CiaoSR with the RDN backbone \cite{zhang2018rdn} can surpass the performance of the better backbone SwinIR \cite{liang2021swinir} (with its original upsampling module).
With the help of the continuous-scale training, a single SR model trained with our method has better generalization performance than vanilla SR models (including EDSR \cite{lim2017edsr}, RDN \cite{zhang2018rdn} and SwinIR \cite{liang2021swinir} trained on specific scale.  
For the out-of-scale experiments, our model also achieves the best generalization performance on unseen scales.
The vanilla EDSR-baseline \cite{lim2017edsr}, RDN \cite{zhang2018rdn} and SwinIR \cite{liang2021swinir} cannot be applied on the out-of-scale experiments because their decoder can only predict the scale that it was trained on. 
We observe that our method yields the largest improvement on Urban100 over existing arbitrary-scale methods (\eg LIIF \cite{chen2021liif} and LTE \cite{lee2022lte}).

\begin{figure*}[t!]
	\centering
	\renewcommand{\g}{-0.7mm}
	\renewcommand{\tabcolsep}{1.8pt}
	\renewcommand{\arraystretch}{1}
	\resizebox{1\linewidth}{!} {
	\hspace{-4.5mm}
	\begin{tabular}{cc}
			\renewcommand{\name}{figures/urban100/004_x4_}
			\renewcommand{\h}{0.12}
			\renewcommand{\w}{0.2}
			\begin{tabular}{cc}
				\begin{adjustbox}{valign=t}
					\begin{tabular}{c}
		         	\includegraphics[trim={0 0 0 0},clip, width=0.354\textwidth]{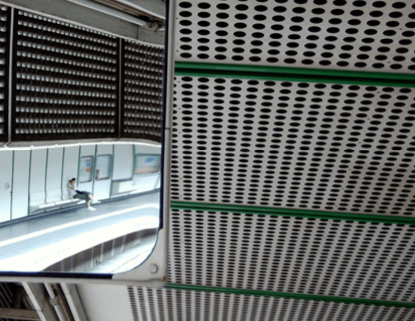}
						\\
						 004, ${\times}4$, Urban100 \cite{Urban100}
					\end{tabular}
				\end{adjustbox}
				\begin{adjustbox}{valign=t}
					\begin{tabular}{cccccc}
						\includegraphics[trim={0 0 0 0},clip,height=\h \textwidth, width=\w \textwidth]{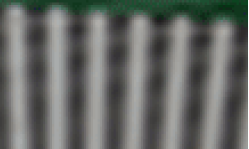} \hspace{\g} &
						\includegraphics[trim={0 0 0 0},clip,height=\h \textwidth, width=\w \textwidth]{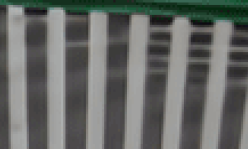} \hspace{\g} &
						\includegraphics[trim={0 0 0 0},clip,height=\h \textwidth, width=\w \textwidth]{\name rdn-metasr.png} & 
						\includegraphics[trim={0 0 0 0},clip,height=\h \textwidth, width=\w \textwidth]{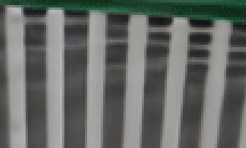} \hspace{\g} 
						\\
						Bicubic \vspace{-0.5pt}\vspace{-4.5pt} & RDN \cite{zhang2018rdn} &
						RDN-MetaSR \cite{hu2019metasr} & RDN-LIIF \cite{chen2021liif} 
						\\
						\vspace{-2mm}
						\\
						\includegraphics[trim={0 0 0 0},clip,height=\h \textwidth, width=\w \textwidth]{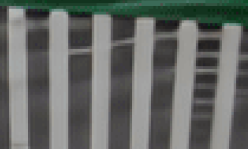} \hspace{\g} &
						\includegraphics[trim={0 0 0 0},clip,height=\h \textwidth, width=\w \textwidth]{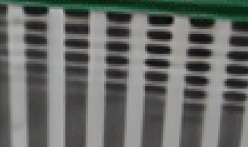} \hspace{\g} &
						\includegraphics[trim={0 0 0 0},clip,height=\h \textwidth, width=\w \textwidth]{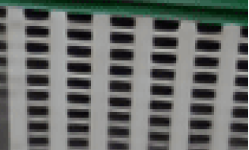}
						\hspace{\g} &		
						\includegraphics[trim={0 0 0 0},clip,height=\h \textwidth, width=\w \textwidth]{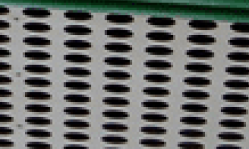}   
						\\ 
						RDN-ITSRN \cite{yang202itsrn} &
						RDN-LTE \cite{lee2022lte}  \hspace{\g} & \textbf{RDN-CiaoSR (ours)} & GT
						\\
					\end{tabular}
				\end{adjustbox}
			\end{tabular}
			
		\end{tabular}
	}
    \resizebox{1\linewidth}{!} {
	\hspace{-4.5mm}
	\begin{tabular}{cc}
			\renewcommand{\name}{figures/manga109/TaiyouNiSmash_x8_}
			\renewcommand{\h}{0.12}
			\renewcommand{\w}{0.2}
			\begin{tabular}{cc}
				\begin{adjustbox}{valign=t}
					\begin{tabular}{c}
		         	\includegraphics[trim={0 0 0 0},clip, width=0.354\textwidth]{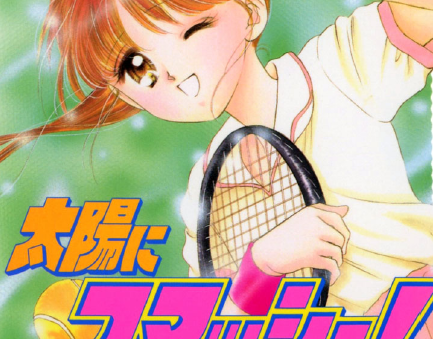}
						\\
						 TaiyouNiSmash, ${\times}8$, Manga109 \cite{Manga109}
					\end{tabular}
				\end{adjustbox}
				\begin{adjustbox}{valign=t}
					\begin{tabular}{cccccc}
						\includegraphics[trim={0 0 0 0},clip,height=\h \textwidth, width=\w \textwidth]{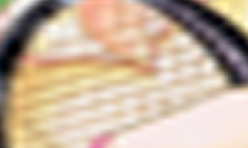} \hspace{\g} &
						\includegraphics[trim={0 0 0 0},clip,height=\h \textwidth, width=\w \textwidth]{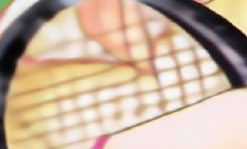} \hspace{\g} &
						\includegraphics[trim={0 0 0 0},clip,height=\h \textwidth, width=\w \textwidth]{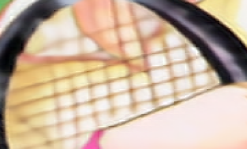} & 
						\includegraphics[trim={0 0 0 0},clip,height=\h \textwidth, width=\w \textwidth]{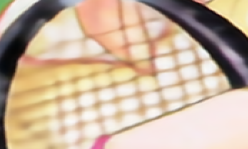} \hspace{\g} 
						\\
						Bicubic \vspace{-0.5pt}\vspace{-4.5pt} & RDN$^*$ \cite{zhang2018rdn} &
						RDN-MetaSR \cite{hu2019metasr} & RDN-LIIF \cite{chen2021liif} 
						\\
						\vspace{-2mm}
						\\
						\includegraphics[trim={0 0 0 0},clip,height=\h \textwidth, width=\w \textwidth]{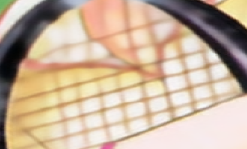} \hspace{\g} &
						\includegraphics[trim={0 0 0 0},clip,height=\h \textwidth, width=\w \textwidth]{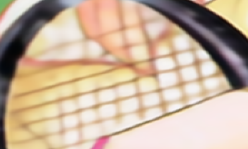} \hspace{\g} &
						\includegraphics[trim={0 0 0 0},clip,height=\h \textwidth, width=\w \textwidth]{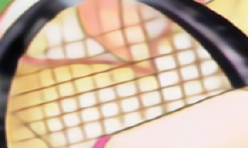}
						\hspace{\g} &		
						\includegraphics[trim={0 0 0 0},clip,height=\h \textwidth, width=\w \textwidth]{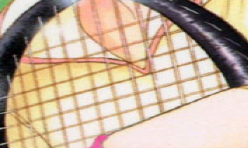}   
						\\ 
						RDN-ITSRN \cite{yang202itsrn} &
						LTE \cite{lee2022lte}  \hspace{\g} & \textbf{RDN-CiaoSR (ours)} & GT
						\\
					\end{tabular}
				\end{adjustbox}
			\end{tabular}
		\end{tabular}
	}
	\vspace{-2mm}
	\caption{Visual comparison of different methods on benchmarks. ``$^*$" means the model first synthesizes twice to $\times8$ images.} 
	\vspace{-2mm}
	\label{fig:sr_quali}
\end{figure*}

\noindent\textbf{Qualitative results.}
In Figure \ref{fig:sr_quali}, we provide qualitative comparisons with other arbitrary-scale SR methods.
Our model is able to synthesize the SR images with sharper textures than other methods.
Taking the second line as an example, the textures in the LR image are degraded, but CiaoSR is still able to restore the textures of the building.
In contrast, other methods can only restore part of the textures possibly due to the limitations of the local ensemble and insufficient features.
More visual results on more test sets are put in the supplementary materials.

\subsection{Ablation Study}
In this section, we conduct ablation studies to investigate effect of each component in our architecture.
Our CiaoSR consists of the attention-in-attention network and a scale-aware non-local attention module.
Table \ref{tab:ablation} shows the contributions of each component on performance.

\noindent\textbf{Attention-in-attention.} The attention-in-attention network is the main branch of our architecture.
On the one hand, the network learns to ensemble weights with both the coordinates and the features. 
On the other hand, the network is able to aggregate the local and non-local information to improve the SR performance.
To investigate how the attention mechanism affects the performance, we remove this module and replace it with an MLP to predict the weights of the ensemble. 
The coordinate information is still fed into the MLP to enable arbitrary scale prediction. 
As shown in Table \ref{tab:ablation}, the method suffers a significant performance drop without the attention mechanism.

\begin{figure*}
  \begin{center}
  \includegraphics[width=1\linewidth]{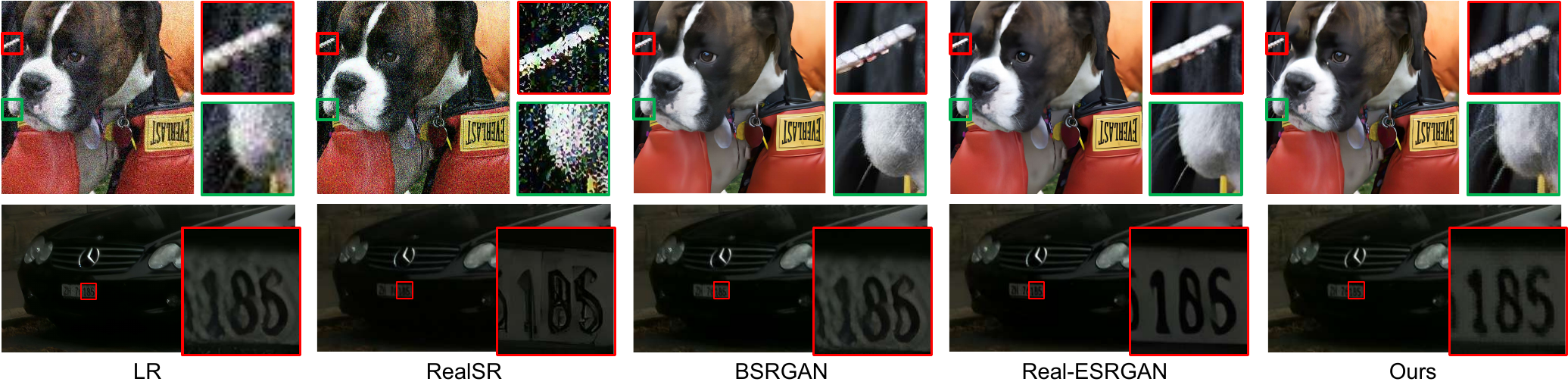}
  \end{center}
  \vspace{-5mm}
  \caption{Visual comparison of different methods on the RealSRSet \cite{kai2021bsrgan} and DPED \cite{ignatov2017dped} dataset (${\times}8$). }
  \label{fig:comp_real}
  \vspace{-5mm}
\end{figure*}

\noindent\textbf{Non-local attention.}
The embedded non-local attention is proposed to capture the scale-aware information from other locations of the images to improve the performance. We investigate the importance of the non-local attention in an implicit model by comparing the results with and without this module. As shown in Table \ref{tab:ablation}, by introducing the non-local attention, a further performance gain is achieved. This is because our module can exploit texture information from other scales in a large perceptual field.

\noindent\textbf{Training with different types of scales.}
We evaluate the effectiveness of our implicit model when trained with discrete (including single scale ($\{2\}$/$\{3\}$/$\{4\}$) and multiple scales $\{2,3,4\}$) and continuous scales $[1, 4]$, and show the results in Table \ref{tab:diff_scale}.
For the discrete type of scales, we observe that training our implicit model with a specific scale can also achieve significant improvement for that specific scale, compared with the original upsampling module in RDN \cite{zhang2018rdn}.
Taking ${\times}2$ scale as an example, our method has an improvement of 0.24dB over RDN.
However, training with a specific scale has poor generalization performance on unseen in-scale and out-of-scale. 
A few existing works~\cite{kim2016vdsr,lim2017edsr} consider the training with multiple discrete scales and found the performance of the joint learned model is \textit{comparable} with single scale networks. 
Based on our implicit attention module, we are able to deal with more scales and exploit the correlation of multiple SR tasks at different scales. Unlike previous works~\cite{kim2016vdsr,lim2017edsr} which only achieve comparable performance compared with single scale models, CiaoSR achieves a better performance than training with one scale.
We found our model has further improvement when trained on continuous scales.
The improvement gain is attributed to our continuous representations design and our architecture, which enables efficient learning from cross-scale information.

\subsection{Further Study}

\vspace{-1mm}
\noindent\textbf{Effect on synthesis steps.}
Our trained model is able to synthesize SR images with any given arbitrary scale in one step. A natural alternative would be to synthesize a large scale in multiple steps. 
Taking a large scale (\eg ${\times}12$) as an example, SR images can be generated by two steps (\eg ${\to}{\times}2{\to}{\times}12$), or three steps (\eg ${\to}{\times}2{\to}{\times}4{\to}{\times}12$).
We compare the performance of synthesizing a large scale in one and multiple steps, and the results on the Urban100 and Manga109 datasets are shown in Table \ref{tab:comp_steps}.
We observe that our way of synthesizing SR images in one step has the best performance under the PSNR and SSIM metrics. 
In contrast, the performance gets worse with more steps because errors can accumulate along the multiple synthesizing steps.

\vspace{1.mm}
\noindent\textbf{Model size and inference time.}
To demonstrate the efficiency of our proposed method, we investigate the model size and inference time of different arbitrary-scale SR methods. 
Results are shown in Table~\ref{tab:modelsize}. 
Specifically, we set the input size as $192{\times}192$ and calculate the inference time on an NVIDIA Quadro RTX 6000.
Our implicit model has much fewer parameters than the previous SOTA methods (\eg LTE \cite{lee2022lte}). 
Our model has the best performance although it requires a little more inference time due to exploiting the non-local features.

\noindent\textbf{More evaluation metrics.}
To further demonstrate the effectiveness of our proposed method, we apply more evaluation metrics (\eg SSIM \cite{wang2004ssim} and LPIPS \cite{zhang2018lpips}) to evaluate the image quality across different methods on the Urban100 dataset.
These evaluation metrics are widely used in the SR community.
In general, higher SSIM and lower LPIPS correspond to better performance.
As shown in Table~\ref{tab:comp_ssim_lpips}, our method achieves the highest SSIM and lowest LPIPS, and thus has the best performance on both in-scale and out-of-scale distributions. 
Compared with other methods, our proposed CiaoSR is able to maintain more structural texture and perceptual information.

\begin{table*}[t]
\begin{minipage}{.45\linewidth}
\centering
\caption{Ablation study on each component of our networks on Urban100. We use RDN \cite{zhang2018rdn} as the backbone.}
\label{tab:ablation}
\resizebox{1\textwidth}{!}{
    \begin{tabular}{l|c||ccc}
    \hline\toprule
    \rowcolor{mygray}
    \multicolumn{2}{l||}{\multirow{-1}{*}{Attention-in-attention}} & \xmark & \cmark & \cmark  \\ 
    \rowcolor{mygray}
    \multicolumn{2}{l||}{\multirow{-1}{*}{Scale-aware Attention Network}} & \xmark & \xmark & \cmark  \\ 
    \hline\hline
    \multirow{3}{*}{In-scale} &
    $\times 2$   & ~~32.87~~ & ~~33.24~~ & ~~\bf{33.30}~~ \\ 
    & $\times 3$ & 28.82 & 29.10 & \bf{29.17} \\
    & $\times 4$ & 26.69 & 26.96 & \bf{27.11} \\
    \hline
    \multirow{2}{*}{Out-of-scale~~~~~~~~~~~} &
    $\times 6$   & 24.22 & 24.50 & \bf{24.58} \\
    & $\times 8$ & 22.80 & 22.98 & \bf{23.13} \\
    \bottomrule
    \end{tabular}
    }
\end{minipage}
~
\begin{minipage}{.55\linewidth}
    \centering
    \caption{Ablation study on training our implicit model with different types of scales on Urban100.}
    \label{tab:diff_scale}
    \resizebox{1\textwidth}{!}{
    \begin{tabular}{l|l||ccc|cc}
    \hline\toprule
    \rowcolor{mygray}
    & & \multicolumn{3}{c|}{In-scale} & \multicolumn{2}{c}{Out-of-scale} \\ 
    \cmidrule{3-7}
    \rowcolor{mygray}
    \multicolumn{1}{l|}{\multirow{-2}{*}{Type}} & \multicolumn{1}{l||}{\multirow{-2}{*}{\!\!Training scale $s$}} & ~~~$\times 2$~~~ & ~~~$\times 3$~~~ & ~~~$\times 4$~~~ & ~~~$\times 6$~~~ & ~~~$\times 8$~~~  \\ 
    \hline\hline
    \multirow{4}{*}{Discrete} &
    $s \in \{2\}$ & 33.13 & 27.01 & 25.60 & 22.27 & 22.09 \\ 
    & $s \in \{3\}$ & 31.39 & 29.06 & 25.77 & 23.44 & 22.16 \\
    & $s \in \{4\}$ & 31.42 & 27.87 & 26.88 & 24.28 & 22.85  \\
    & $s \in \{2, 3, 4\}$~~~~ & 33.15 & 29.14 & 27.02 & 24.47 & 23.03  \\
    \hline
    Continuous~~~~ & $s\in[1, 4]$ & \bf{33.30} & \bf{29.17} & \bf{27.11} & \bf{24.58} & \bf{23.13} \\ 
    \bottomrule
    \end{tabular}
    }
    \end{minipage}
    \vspace{-3mm}
\end{table*}

\begin{table*}[t]
\begin{minipage}{.47\linewidth}
\centering
\caption{Comparison of (PSNR and SSIM) for different synthesis steps on Urban100 \cite{Urban100} and Manga109 \cite{Manga109}.}
\label{tab:comp_steps}
\resizebox{1\textwidth}{!}{
\begin{tabular}{l|l||c|c|c|c}
\hline\toprule
\rowcolor{mygray}
& &\multicolumn{2}{c|}{Urban100 \cite{Urban100}} & \multicolumn{2}{c}{Manga109 \cite{Manga109}}  \\
\cmidrule{3-6}
\rowcolor{mygray}
\multirow{-2}{*}{Type} &\multirow{-2}{*}{Synthesis steps}
& \multicolumn{1}{c|}{PSNR} & \multicolumn{1}{c|}{SSIM} & \multicolumn{1}{c|}{PSNR} & \multicolumn{1}{c}{SSIM}  \\
\hline\hline
\multirow{2}{*}{Multiple steps}
&${\to}{\times}2{\to}{\times}4{\to}{\times}12$ &21.28 & 0.557 & 22.46 & 0.720 \\
&${\to}{\times}2{\to}{\times}12$ & 21.32 & 0.558 & 22.53 & 0.721  \\
One step & ${\to}{\times}12$ & \bf{21.42} & \bf{0.561} & \bf{22.63} & \bf{0.723}  \\
\bottomrule
\end{tabular}
}
\end{minipage}
~
\begin{minipage}{.53\linewidth}
\centering
\caption{Comparisons of model size, inference time and performance gain of different models.}
\label{tab:modelsize}
\resizebox{1\textwidth}{!}{
\begin{tabular}{l||ccccc}
\hline\toprule
\rowcolor{mygray}
Different models & \!\!Meta-SR \cite{hu2019metasr}\!\! & \!\!LIIF \cite{chen2021liif}\!\! & \!\!ITSRN \cite{yang202itsrn}\!\! & \!\!LTE \cite{lee2022lte}\!\! & CiaoSR \\ 
\hline\hline
Model size (M)   & 1.7 & 1.6 & 0.7 & 1.7 & 1.4 \\
Inference time (ms)  & 237 & 171 & 343 & 148 & 528 \\
PSNR (dB) & 26.55 & 26.68 & 26.77 & 26.81 & 27.11 \\
\hline
Performance gain (dB) & -0.06 & 0.07 & 0.16 & 0.2 & \bf{0.5} \\
\hline
\end{tabular}
}
\end{minipage}
\vspace{-3mm}
\end{table*}

\begin{table*}[h!]
    \begin{minipage}{.62\linewidth}
    \centering
    \caption{Comparison of more evaluation metrics (SSIM$\uparrow$ \cite{wang2004ssim} and LPIPS$\downarrow$ \cite{zhang2018lpips}) for different methods on Urban100.}
    \label{tab:comp_ssim_lpips}
    \resizebox{1.0\textwidth}{!}{
    \begin{tabular}{l||cc|cc|cc|cc|cc}
    \hline\toprule
    \rowcolor{mygray}
    & \multicolumn{6}{c|}{In-scale} & \multicolumn{4}{c}{Out-of-scale} \\ 
    \cmidrule{2-11} 
    \rowcolor{mygray}
    \multirow{-1}{*}{\!\!RDN-Method\!\!}
    & \multicolumn{2}{c|}{$\times 2$} & \multicolumn{2}{c|}{$\times 3$} & \multicolumn{2}{c|}{$\times 4$} & \multicolumn{2}{c|}{$\times 6$} & \multicolumn{2}{c}{$\times 8$}  \\
    \cmidrule{2-11}
    \rowcolor{mygray}
    & \multicolumn{1}{c}{\!\!SSIM\!\!} & \multicolumn{1}{c|}{\!\!LPIPS\!\!} & \multicolumn{1}{c}{\!\!SSIM\!\!} & \multicolumn{1}{c|}{\!\!LPIPS\!\!} & \multicolumn{1}{c}{\!\!SSIM\!\!} & \multicolumn{1}{c|}{\!\!LPIPS\!\!} & \multicolumn{1}{c}{\!\!SSIM\!\!} & \multicolumn{1}{c|}{\!\!LPIPS\!\!} & \multicolumn{1}{c}{\!\!SSIM\!\!} & \!\!LPIPS\!\! \\
    \hline\hline
    \!\!LIIF \cite{chen2021liif} & \!\!0.934\!\! & \!\!0.104\!\! & \!\!0.866\!\! & \!\!0.197\!\! & \!\!0.804\!\! & \!\!0.267\!\! & \!\!0.704\!\! & \!\!0.360\!\! & \!\!0.636\!\! & \!\!0.425\!\! \\
    \!\!MetaSR \cite{hu2019metasr}\!\!\! & \!\!0.935\!\! & \!\!0.104\!\! & \!\!0.866\!\! & \!\!0.195\!\! & \!\!0.805\!\! & \!\!0.261\!\! & \!\!0.698\!\! & \!\!0.353\!\! & \!\!0.624\!\! & \!\!0.423\!\! \\
    \!\!ITSRN \cite{yang202itsrn} & \!\!0.936\!\! & \!\!0.102\!\! & \!\!0.868\!\! & \!\!0.193\!\! & \!\!0.807\!\! & \!\!0.259\!\! & \!\!0.703\!\! & \!\!0.350\!\! & \!\!0.633\!\! & \!\!0.417\!\! \\
    \!\!LTE \cite{lee2022lte} & \!\!0.936\!\! & \!\!0.104\!\! & \!\!0.868\!\! & \!\!0.195\!\! & \!\!0.807\!\! & \!\!0.264\!\! & \!\!0.707\!\! & \!\!0.356\!\! & \!\!0.639\!\! & \!\!0.421\!\! \\
    \!\!\textbf{CiaoSR}\!\! & \!\!\bf{0.938}\!\! & \!\!\bf{0.100}\!\! & \!\!\bf{0.871}\!\! & \!\!\bf{0.188}\!\! & \!\!\bf{0.814}\!\! & \!\!\bf{0.255}\!\! & \!\!\bf{0.718}\!\! & \!\!\bf{0.346}\!\! & \!\!\bf{0.651}\!\! & \!\!\bf{0.408}\!\! \\ 
    \bottomrule
    \end{tabular}
    }
  \end{minipage}
  ~
  \begin{minipage}{.38\linewidth}
    \centering
    \caption{Quantitative comparisons of different Real-world SR methods on RealSRSet \cite{kai2021bsrgan}.}
    \label{tab:comp_real}
    \resizebox{1.0\textwidth}{!}{
    \begin{tabular}{l|c||ccc}
    \hline\toprule
    \rowcolor{mygray}
    \!\!Methods      & \!\!Scale\!\! & NIQE$\downarrow$ & \!\!\!\!BRISQUE$\downarrow$\!\!\!\! & PIQE$\downarrow$  \\
    \hline\hline
    \multirow{2}{*}{RealSR \cite{cai2019realsr}} 
    & ${\times}$8  & 4.7949	& 23.3228 & 29.3976 \\
    & ${\times}$16 & 3.3946	& 21.6390 & 21.4116 \\ 
    \hline
    \multirow{2}{*}{BSRGAN \cite{kai2021bsrgan}} 
    & ${\times}$8  & 4.7007 & 37.0638 & 36.1889 \\
    & ${\times}$16 & 4.1408 & 39.6252 & 37.9176 \\ 
    \hline
    \multirow{2}{*}{Real-ESRGAN \cite{wang2021realesrgan}\!\!} 
    & ${\times}$8  & 4.5280 & 40.6521 & 59.8915 \\
    & ${\times}$16 & 5.6442 & 53.9364 & 84.3611 \\ 
    \hline
    \multirow{2}{*}{Ours-real} 
    & ${\times}$8  & 3.5894 & 30.9219 & 29.8554 \\
    & ${\times}$16 & 3.9461 & 41.8145 & 58.8244 \\ 
    \hline
    \end{tabular}
    }
  \end{minipage}
  \vspace{-1mm}
\end{table*}

\subsection{Real-World Arbitrary-Scale Image SR}
\vspace{-2mm}
We extend our proposed method to real-world applications with the ultimate goal of synthesizing real images with arbitrary scales. 
Specifically, we train our model on the DIV2K dataset \cite{DIV2K}, and test on the RealSRSet \cite{kai2021bsrgan} and DPED \cite{ignatov2017dped} datasets.
We apply the practical degradation models (including blur, noise, compression, etc.) of BSRGAN \cite{kai2021bsrgan} and Real-ESRGAN \cite{wang2021realesrgan} to synthesize paired training data on DIV2K \cite{DIV2K}.
We set the HR patch size of training data as 256, and the batch size as 48. We use Adam \cite{kingma2014adam} as the optimizer, and adopt the exponential moving average (EMA). We first train our model with L1 loss for 1000K iterations with the learning rate of $2{\times}10^{-4}$ and train with L1 loss, perceptual loss and GAN loss together for 400K iterations with the learning rate of $1{\times}10^{-4}$. More training details can be found in the supplementary materials.

Since there are no arbitrary-scale SR methods applied to real-world images, we mainly compare with the state-of-the-art real-world image SR models, including RealSR \cite{cai2019realsr}, BSRGAN \cite{kai2021bsrgan} and Real-ESRGAN \cite{wang2021realesrgan}.
Because there are no ground-truth (GT) real images, we apply non-reference image quality assessment (\eg NIQE \cite{mittal2012niqe}, BRISQUE \cite{mittal2011BRISQUE} and PIQE \cite{venkatanath2015PIQE}) to evaluate the image quality. 
As shown in Table \ref{tab:comp_real}, our model does not optimize for all metrics since these metrics cannot well match the actual human perceptual \cite{blau2018pi}.
To address this, we further compare the visual results on the RealSRSet \cite{kai2021bsrgan} and DPED \cite{ignatov2017dped} datasets in Figure \ref{fig:comp_real}.
Our proposed method is able to synthesize real SR images with realistic textures.
In particular, our method is able to synthesize more details of a dog in the first line and more clear digit `5' in the second line, compared with other methods.
In contrast, other methods often introduce unnatural artifacts within images, especially for out-of-scale distributions.
To further investigate the effectiveness of our model, we provide more visual results of our real model in the supplementary materials.

\section{Conclusion}
\label{sec:conclusion}
\vspace{-2mm}
In this paper, we have proposed a novel continuous implicit attention-in-attention network for arbitrary-scale image super-resolution, called CiaoSR.
Specifically, our CiaoSR first introduces a scale-aware non-local attention in our attention to exploit more relevant features, and then
predicts RGB values by locally ensembling features with the learnable attention weights.   
More importantly, CiaoSR has good flexibility and applicability since it can be used behind any SR backbone to boost the performance.
Extensive experiments on all benchmark datasets demonstrate the superiority of our CiaoSR in both SISR and arbitrary-scale SR tasks.
Besides, our CiaoSR has good generalization performance on both in-scale and out-of-scale distributions.
Last, we extend our method to the real-world application to synthesize real SR images with arbitrary scale.

\vspace{1mm}
\noindent\textbf{Acknowledgements:}
This work was partly supported by the Huawei Fund, the ETH Z\"urich General Fund (OK) and the Alexander von Humboldt Foundation.

{\small
\bibliographystyle{ieee_fullname}
\bibliography{egbib}
}

\end{document}